\documentclass{article}

\usepackage{arxiv}

\usepackage[utf8]{inputenc} 
\usepackage[T1]{fontenc}    
\usepackage{hyperref}       
\usepackage{url}            
\usepackage{booktabs}       
\usepackage{amsfonts}       
\usepackage{nicefrac}       
\usepackage{microtype}      
\usepackage{lipsum}
\usepackage{graphicx}
\graphicspath{ {./images/} }

\usepackage{caption}
\usepackage{algorithm,algcompatible,amsmath} 
\usepackage{etoolbox}
\usepackage{tikz}
\usepackage{mathtools, cuted}
\usepackage{yfonts}
\usepackage{txfonts}
\usepackage{arydshln}
\usepackage{tabularx}
\usepackage{mwe}

\DeclareMathOperator*{\argmax}{\arg\!\max}
\DeclareMathOperator*{\argmin}{\arg\!\min}
\algnewcommand\INPUT{\item[\textbf{Input:}]}%
\algnewcommand\OUTPUT{\item[\textbf{Output:}]}%

\makeatletter
\newcommand{\algorithmicbreak}{\textbf{break}}
\newcommand{\BREAK}{\STATE \algorithmicbreak}
\expandafter\patchcmd\csname\string\algorithmic\endcsname{\itemsep\z@}{\itemsep=0.7ex plus2pt}{}{}
\makeatother

\hypersetup{
  colorlinks=true,
  linkcolor=blue!50!red,
  urlcolor=blue!70!black,
  citecolor=blue
}

\newcommand*{\bigCI}{%
  \mathrel{\text{%
    {\begin{tikzpicture}[baseline = {(0,0)}]%
        \draw[line width = 0.1ex] (0,0) -- (0,1.7ex);
        \draw[line width = 0.1ex] (0.5ex,0) -- (0.5ex,1.7ex);
        \draw[line width = 0.1ex] (-0.8ex,0) -- (1.3ex,0);
     \end{tikzpicture}%
    }%
  }}%
}

\DeclareMathOperator{\EX}{\mathbb{E}}
\newcommand{\vt}[1]{\boldsymbol{#1}}                
\newcommand*{\MyIndent}{\hspace*{0.5cm}}
\makeatletter
\def\ps@pprintTitle{%
   \let\@oddhead\@empty
   \let\@evenhead\@empty
   \let\@oddfoot\@empty
   \let\@evenfoot\@oddfoot
}
\makeatother

\title{Learning medical triage from clinicians using Deep Q-Learning}

\author{Albert Buchard
        \And
        Baptiste Bouvier
        \And
        Giulia Prando
        \And
        Rory Beard
        \And
        Michail Livieratos
        \And
        Dan Busbridge
        \And
        Daniel Thompson
        \And
        Jonathan Richens
        \And
        Yuanzhao Zhang
        \And
        Adam Baker
        \And
        Yura Perov
        \And
        Kostis Gourgoulias
        \And
        Saurabh Johri 
        \And 
        \hspace{350pt}  \\ 
        Babylon Health
        }
 
\date{}

\begin{document}
\maketitle
\begin{abstract}

Medical Triage is of paramount importance to healthcare systems, allowing for the correct orientation of patients and allocation of the necessary resources to treat them adequately. While reliable decision-tree methods exist to triage patients based on their presentation, those trees implicitly require human inference and are not immediately applicable in a fully automated setting. On the other hand, learning triage policies directly from experts may correct for some of the limitations of hard-coded decision-trees. In this work, we present a Deep Reinforcement Learning approach (a variant of Deep Q-Learning) to triage patients using curated clinical vignettes. The dataset, consisting of 1374 clinical vignettes, was created by medical doctors to represent real-life cases. Each vignette is associated with an average of $3.8$ expert triage decisions given by medical doctors relying solely on medical history. We show that this approach is on a par with human performance, yielding safe triage decisions in 94\% of cases, and matching expert decisions in 85\% of cases. The trained agent learns when to stop asking questions,  acquires optimized decision policies requiring less evidence than supervised approaches, and adapts to the novelty of a situation by asking for more information. Overall, we demonstrate that a Deep Reinforcement Learning approach can learn effective medical triage policies directly from clinicians’ decisions, without requiring expert knowledge engineering. This approach is scalable and can be deployed in healthcare settings or geographical regions with distinct triage specifications, or where trained experts are scarce, to improve decision making in the early stage of care.

\end{abstract}


\section{Introduction}
\subsection{Medical Triage}
For many patients, medical triage is the first organising contact with the healthcare system. Be it through a telephone interaction, or performed face-to-face by a trained healthcare professional, the triage process aims to uncover enough medical evidence to make an informed decision about the appropriate point of care given a patient's presentation. The clinician's task is to plan the most efficient sequence of questions in order to make a fast and accurate triage decision. 
Although internationally recognized systems exist \cite{Eitel2003, Cronin2003,Beveridge1999, Gottschalk2006, VanIerland2011, Ebrahimi2015}, with clearly defined decision-trees based on expert consensus,  in practice, the nature of the triage task is not a  passive recitation of a  learned list of questions.  Triage is an active process through which the clinician must make inferences about the causes of the patient's presentation and update his plan following each new piece of information.

To deploy triage systems in healthcare settings, a population of clinicians needs to go through training to ensure the reliability and quality of their practice. No triage system seems to be superior overall, but their performance varies significantly across studies \cite{Zachariasse2019}. In order to improve patient safety and quality of care, many decision-support tools were designed on top of those decision-trees to standardize and automate the triage process (eCTAS\cite{McLeod2019}, NHS111\cite{Turner2013}) with mixed results. 

At the turn of the last decade, the field of automated clinical decision-making was invigorated by the revolution in deep-learning approaches with several applications in perceptual settings, in which the decision relies on image recognition and not on clinical signs \cite{Annarumma2019,McKinney2020}. However, research on automated triage systems that do not rely on expert-crafted decision trees, and able to learn from data, is still relatively sparse. This may be due to the meagre availability of detailed and clean Electronic Health Records and the fact that triage systems rely on highly symbolic and structured inputs, such as symptoms and physical signs.

Ideally, learning a triage system from a detailed distribution of patients' clinical history would allow us to correct for the inherent biases of expert-crafted systems and tailor the system for a specific target population. Given enough data, we could train such systems to improve healthcare outcomes directly and enhance clinical decision making at the early stage of care to minimize the risk for patients.

A more straightforward approach is to learn medical triage policies directly from expert decisions. Using judgments from medical experts made over a dataset of patient presentations, we may hope to learn effective policies which reflect guidelines and the combined experience of many clinicians.  Our work presents a Reinforcement Learning approach for learning medical triage directly from expert decisions. 

\subsection{Reinforcement Learning}
Reinforcement Learning (RL) is a natural approach for problems requiring the optimization of sequences of actions in order to reach complex objectives. Although interaction with real patients is usually not ethically possible, researchers have used observational and generated datasets to apply RL approaches to the healthcare setting. Those methods have been successfully applied to solve complex tasks, such as treatment regimes optimization, precision medicine, automated diagnosis, and personal health assistance (for a recent review, see \cite{Yu}). 

RL describes an approach to learning where an agent learns through interactions with an environment, gathering rewards and penalties for the actions it performs. Under the paradigm of RL, the general interaction between the agent and its environment is well defined.

The environment describes the world in which the agent is evolving in time. At each time $t$:
\begin{itemize}
    \item It keeps track of the agent's state ${s_{t} \in \mathcal{S}}$, with $\mathcal{S}$ referring to the state-space, i.e. the set of all valid states the agent can be in. 
    \item It processes the agent's actions $a_t \in \mathcal{A}$, with $\mathcal{A}$ the set of all possible actions called the action-space of the agent.
    \item It encodes the system dynamics, fully defined by the transition function ${p: \mathcal{S} \times \mathcal{A} \times \mathcal{S} \rightarrow [0,1]}$ which gives the probability to transition to  state ${s'}$ given that the agent performed action $a$ in state $s$:
    $$p\left(s'\bigm\vert s,a\right) = P\left(s_{t+1}=s' \bigm\vert s_t=s, a_t=a\right).$$
    \item It defines a notion of optimal behavior through a reward function ${r : \mathcal{S} \times \mathcal{A} \rightarrow \mathbb{R}}$, here defined as a map from a state-action pair to a real value (e.g +1 for a positive reward, and -1 for a penalty), which is returned at each time step to the agent.
\end{itemize}

An agent is an entity that performs actions into the environment, given its current state and a policy ${\pi: \mathcal{S} \times \mathcal{A} \rightarrow [0,1]}$, a function that gives the probability of an action when in a particular state. The set of actions that an agent takes and the set of states that it consequently visits constitute a trajectory $\tau\in\mathcal{T}$, denoted as $\tau=(s_0, a_0, s_1, a_1, ...)$. The goal of training an RL agent is to learn the optimal policy $\pi* : \mathcal{S} \rightarrow \mathcal{A}$, a deterministic function that maps an agent's state to a specific action so that the reward received by the agent is maximized in expectation across all interactions. 

\subsection{Q-Learning}
\emph{Model-free} RL algorithms deal with settings where an agent does not have access to the dynamics of the environment and cannot interrogate or learn the transition function. Two main classes of model-free algorithms exist: a) policy-based methods, which aim to learn the policy directly, and b) value-based methods which aim to learn one or several \emph{value functions} to guide the agent policy toward high reward trajectories. In this work, we will use a variant of \emph{Q-Learning} \cite{Watkins1992}, which is a model-free and value-based method, called Deep Q-Learning \cite{Mnih}. 

In Q-Learning, the agent does not learn a policy function directly but instead learns a proxy value-function $Q(s, a)$. This function approximates an optimal function $Q^*(s, a)$ defined as the maximum expected return achievable by any policy $\pi$, over all possible trajectories $\tau$, given that in state $s$ the agent performs action $a$ and the rest of the trajectory $\tau$ is generated by following $\pi$, denoted as $\tau\sim\pi$ (with a slight abuse of notation).
\begin{equation}
    Q^*(s,a) = \max_{\pi} \mathbb{E}_{\tau \sim \pi} \left[ R\left(\tau \big\vert s_0 = s, a_0 = a\right) \right],
\end{equation} 
$R(\tau| s_0, a_0)$ is the function returning the reward gathered over the trajectory $\tau$ defined as:
\begin{equation}
    R\left(\tau \bigm\vert s_0, a_0\right) = r\left(s_0, a_0\right) + \overset{\infty}{\underset{s_t, a_t \in \tau}{\sum}} \gamma^t r\left(s_t, a_t\right)
\end{equation} 
The weight $\gamma \in \left[0, 1\right]$ appearing in the previous equation is called the \emph{discount} factor. It encodes the notion that sequences of actions are usually finite, and one should give more weight to the current reward. $Q^*(s,a)$ has an optimal substructure and can be written recursively using the Bellman Equation \cite{Bellman1957}, which treats each decision step as a separate sub-problem:
\begin{equation}
    Q^*(s,a) = r(s,a) + \gamma \sum_{s'} p(s' | s, a) \max_{a'} Q^*(s', a').
\end{equation} 
This function encodes the value of performing a particular action $a$ when in state $s$ as the sum of the immediate reward returned by the environment and the weighted expected rewards obtained over the future trajectory generated by a greedy policy. 
 
During Q-Learning, experience tuples  ${e_i := \left(s, a, r, s' \right)}$ of the agent's interaction with its environment are usually stored in a memory $\mathcal{M}$; each record is composed of an initial state $s$, the chosen action $a$, the received reward $r$ and next state $s'$. During learning the agent samples records from past experiences and learns the optimal Q-value function by minimizing the \emph{temporal difference error} (TD-Error), defined as the difference between a \emph{target Q-value} computed from a record ${e_i}$ and the current Q-value  for a particular state-action pair $(s,a) \in e_i$:
\begin{equation}
    Q^*(s,a) = \underset{Q}{\argmin}\ \mathbb{E}_{\substack{e_i \sim \mathcal{M}\\ s,a \in e_i}} \left[
    Q^T\left(s, a \bigm\vert e_i\right) - Q\left( s,a \right)\right].
\end{equation}  
 The \emph{target Q-value} ($Q^T$) is computed from an experience tuple $e_i$ by combining the actual observed reward, and the maximum future expected reward:
\begin{equation}
   Q^T\left(s, a \bigm\vert e_i\right) = \left\{ \begin{array}{ll} r  & \mbox{if } a \mbox{ is terminal,}\\
   r + \gamma\ \underset{a'}{\max} Q\left(s',a'\right) & \mbox{otherwise.}
   \end{array}\right.
\end{equation} 
In practice, the Q-values are updated iteratively from point samples until convergence or until a maximum number of steps is completed. At each iteration, the new Q-value for the state-action pair $(s,a)\in e_i$ is then defined as
\begin{equation}
    {Q(s,a) \gets  (1 - \alpha) Q(s,a) + \alpha  Q^T\left(s, a \bigm\vert e_i\right) },
\end{equation} 
with $\alpha$ the learning rate of the agent.

Notice that this algorithm requires the value of $Q$ for each state-action pair $(s,a)$ to be stored somehow. Hence, the classic Q-Learning algorithm and other tabular RL methods often fall short in settings with large state-action spaces, which strongly constrain its potential use in healthcare. For example, in our case, the state space has $|\mathcal{S}| = 3^{597}$ possible configurations, corresponding to the 597 elements of the set of observable medical evidence $\mathcal{E}$ (symptom or risk factor). The set $\mathcal{E}$ corresponds to a subset of the clinical evidences used by the PGM model in production at Babylon Health, each of which is in one of three states: unobserved, observed present, or observed absent.

\subsection{Deep Q-Learning}
Deep Reinforcement Learning refers to a series of new reinforcement learning algorithms that employ (Deep) Neural Networks (NNs) to approximate essential functions used by the agent. NNs amortize the cost of managing sizeable state-action spaces, both in terms of memory and computation time, and are able to learn complex non-linear functions of the state. They are used in particular to learn a policy function directly or to learn a value function.
Deep RL is better suited to handle the complex state-space associated with healthcare-related tasks. Those tasks often require reasoning over large state spaces of structured inputs composed of healthcare events, medical symptoms, physical signs, lab tests, or imagery results. 

Deep Q-Learning (DQN) is a now-famous approach, which uses a NN to learn the Q-value of the state-action pairs $Q_{\theta}(s, a)$, with $\theta$ the parameters of the network. The core of the approach remains similar to classic Q-Learning but now uses stochastic gradient descent, rather than an explicit tabular update, to update $\theta$ following the gradient that minimizes the squared TD-error for each batch $\xi_j \subset \mathcal{M}$:
\begin{equation}
\label{eq:sgd-q-update}
    \begin{matrix*}[l]
    \mathcal{L}(\theta_j) & = & \mathbb{E}_{\substack{e_i \sim \xi_j\\s, a \in e_i}} \left[ \left( Q_{\theta_j}^T\left(s, a \bigm\vert e_i\right) - Q_{\theta_j}(s, a) \right)^2 \right]\\
    \\
    \theta_{j+1} & = & \theta_{j} - \alpha \Delta_{\theta_j} \mathcal{L}(\theta_j).
    \end{matrix*}
\end{equation}

\leavevmode\par
In this study, we present a deep reinforcement learning approach to medical triage, where an artificial agent learns an optimized policy based on expert-crafted clinical vignettes. The trained agent matched human performance with an appropriate triage decision of $85\%$ on previously unseen cases, and compared to a purely supervised method, has the advantage of learning a compressed policy by learning when to stop asking questions. We do not train the agent to ask specific questions, and our approach can be used in conjunction with any question-asking system, be it human, rule-based, or model-based \cite{Buchard2019}. Future work on active triage will concentrate on training an agent to solve both the question-asking and triage tasks simultaneously.

\section{Methods}
\subsection{Clinical vignettes}
The training and testing of the model relied upon a dataset $\mathcal{D}$ composed of $1374$ clinical vignettes each describing a patient presentation  ${V_i := \left\{ v_k \mid v_k \in \mathcal{E}^{+/-}  \right\}}$ containing elements from the set   $\mathcal{E}^{+/-}$ of all potential clinical evidences. Each $v_k$ represents a symptom or a risk factor, known to be either absent or present, with  ${|\mathcal{E}^{+/-}| = 1194}$. Each vignette $V_i$ is associated with an average of 3.36 ($SD=1.44$) expert triage decisions
$${A_i = \left\{ a_j^{m\left(a_j\right)} \bigm\vert a_j \in \mathcal{A} \right\}},$$
with ${\mathcal{A} := \left\{ Red, Yellow, Green, Blue \right \}}$ here denoting the set of valid triage decisions, and $m\left(a_j\right) \in \mathbb{N}$ the multiplicity of the decision $a_j$ in the multiset $A_i$ (see Table \ref{tab:vignettes_stats}). This four-colour based system indicates how urgently a patient should be seen. It is similar to the one used by the Manchester Triage Group (MTG) telephone triage \cite{ManchesterTriageGroup2015}, which simplifies to four categories the widely used 5-colour triage system in the triage literature. Red is associated with life-threatening situations, which requires immediate attention. Yellow indicates that the patient should be seen within the next couple of hours, Green that the patient should be seen but not urgently, and Blue that the patient should be given self-care advice and be directed towards a pharmacy if necessary. The validity of each vignette was evaluated independently by two clinicians. The triage decisions associated with each vignette were made by separate clinicians, blinded to the true underlying disease of the presentation. Critically, the clinician's triage policy, which we aim to learn, was left to their expertise and was not constrained by a known triage system like the MTG.

\begin{figure}[ht]
\centering
\vspace{0pt}
\begin{tabular}{l c c}
\toprule
  & Mean (SD) & N (Vignettes) \\ 
\cmidrule(lr){2-3} 
\textbf{Evidences} & & 1374\\ 
\MyIndent \# Symptoms & 6.15 (2.6) &  \\
\MyIndent\MyIndent \textit{Present} & 3.70 (1.85) &  \\
\MyIndent\MyIndent \textit{Absent} & 2.45 (2.14) &  \\
\MyIndent \# Risk factors & 1.13 (1.14) &  \\
\MyIndent\MyIndent \textit{Present} & .79 (.89) &  \\
\MyIndent\MyIndent \textit{Absent} & .34 (.72) &  \\ 
\textbf{Triage decisions} & & 1374\\ 
\MyIndent Number per vignette & 3.36 (1.44) &  \\
\MyIndent Distribution &  &  \\
\MyIndent\MyIndent \textit{Red} & .09 (.24) &  \\
\MyIndent\MyIndent \textit{Yellow} & .34 (.36) &  \\
\MyIndent\MyIndent \textit{Green} & .48 (.37) &  \\
\MyIndent\MyIndent  \textit{Blue} & .09 (.22) &  \\
\textbf{Inter-expert metrics} & & 1073 \\
\MyIndent Appropriateness & .84 (.17) &  \\
\MyIndent Safety & .93 (.12) &  \\
\bottomrule
\end{tabular}
\captionof{table}{Clinical vignettes statistics. The clinical vignettes ($N=1374$) are a sparse representation of a patient presentations, with an average of $6.15\ (sd=2.5)$ symptoms and $1.13\ (sd=1.14)$ risk factors per vignette. They are associated with expert triage decisions from an average of $3.36\  (sd=1.44)$ independent clinicians ($Min=1$, $Max=11$) having access to all the evidence on the vignette to make their decision. The inter-expert appropriateness and safety is a proxy for evaluating the average expert's performance (see Section \ref{baselines}), and is only evaluated on vignettes with at least three different triage decisions ($N=1073$).}
\label{tab:vignettes_stats}
\end{figure}

\subsection{The state-action space}
In the task we are considering, at each time step the agent performs one of the five available actions ${\mathcal{A}^+ := \mathcal{A} \bigcup ask}$. That is, it either asks for more information, or it makes one of the four triage decisions. 
For each vignette, the set of medical evidence $V_i$ is mapped to a full state vector representation ${E_i \in \mathcal{S}}$, with ${\mathcal{S} := \left\{ -1, 0, 1 \right\}^{\mid \mathcal{E} \mid}}$ the state-space. Specifically, each element takes value $-1$ if the corresponding symptom or risk factor is known to be absent (e.g. absence of fever), $+1$ for known positive evidence (e.g., headache present), or $0$ for unobserved evidence.    

\subsection{The Vignette environment}
At each new episode, the environment is configured with a new clinical vignette. The environment processes the evidence and triage decisions on the vignette and returns an initial state $s_0$ with only one piece of evidence revealed to the agent, i.e. $s_0$ is a vector of all zeroes of size $|\mathcal{E}|$ except for one element which is either $+1$ or $-1$. At each time step $t$, the environment receives an action $a_t$ from the agent. If the agent picks one of the four triage actions, the episode ends, and the agent receives a final reward (see Section \ref{Counterfactual Reward} for more detail on the reward shaping). If the agent asks for more evidence, the environment uniformly samples one of the missing pieces of evidence and adds it to the state $s_{t+1}$. During training, we force the agent to make a triage decision if no more evidence is available on the vignette.

\subsection{The Agent}
The agent architecture follows a DQN approach \cite{Mnih}. The network is composed of four fully connected layers. The input layer takes the state vector $s_t  \in \left\{ -1, 0, 1 \right\}^{\mid \mathcal{E} \mid}$, the hidden layers are fully connected layers with 1024 scaled exponential linear units \cite{Klambauer2017}, and the output layer $l_{out} \in [0,1]^{\vert \mathcal{A}^+ \vert}$ uses a sigmoid activation function.  Keeping $l_{out}$ between 0 and 1 allows easier reward shaping: by limiting the valid range for the rewards and treating them as probabilities of being the optimal action, rather than arbitrary scalar values. Observations gathered by the agent are stored in a variant of the Prioritised Experience Replay Memory (PER) \cite{Schaul2015} and replayed in batches of 100 independent steps during optimisation. After a burn-in period of 1000 steps during which no learning occurs, we then train the agent on a randomly sampled batch after each action. To promote exploration during training, instead of using a classic $\epsilon$-greedy approach, a small amount of Gaussian noise ${\epsilon \sim \mathcal{N}\left( 0,\ \sigma\left(t\right)\right)}$ is added to $Q\left(s_t, ask  \right)$ before the greedy policy picks the action with the highest Q-value: 
\begin{equation}
    a_t = \underset{a}{\argmax}\ Q_{\theta}(s_t, a) + \left[a = ask\right] \mathcal{N}\left(0, \sigma\left(i\right)\right),
\end{equation} 
where the operator $[a=ask]$ is the Iverson bracket, which converts any logical proposition into a number that is 1 if the proposition is satisfied, and 0 otherwise. The noise standard deviation $\sigma(t)$ is decayed from ${\sigma(t_0)=0.05}$ initially to ${\sigma(t_{>3000})=0.001}$ (see Algorithm \ref{algo:dyqn}). 

The noise is only added to action \textit{ask} and not to the triage actions $\mathcal{A}$ because the goal of exploration is to evaluate when to stop rather than to gather information about specific triage rewards. Here the triage actions are terminal, and all receive a \emph{counterfactual reward}, which is independent of the action performed at each time step.
 
\subsection{Counterfactual reward}
\label{Counterfactual Reward} 
One key difference with other RL settings is that the rewards are not delayed, and akin to a supervised approach, each action receives a reward, whether the agent chose that action or not. At each time step, the reward received by the agent is then not a scalar, but a vector $\vt{r} \in [0,1]^{\vert \mathcal{A} \vert}$ which represents the reward for each of the four triage actions. The \emph{ask} action does not receive a reward from the environment (see Section \ref{dynamic_qn}). The reward informs all the agent's actions rather than only the single-action it selected, \emph{as if} it had done all actions at the same time in separate counterfactual worlds. 

Reward shaping was of crucial importance for this task, and many reward schemes were tested to fairly promote the success metrics of Appropriateness and Safety (see Section \ref{metrics}). Trying to balance their relative importance in the reward proves to be less efficient than trying to match the distribution of experts' triage decisions. Hence for every vignette $V_i$, each triage decision $a\in\mathcal{A}$ is mapped to a reward equal to the normalised probability of that decision in the bag of expert decisions $A_i$. Namely, denoting the element of $\vt{r}$ corresponding to the reward for action $a$ as $\vt{r}_a$, we define:
\begin{equation} 
  \vt{r}_a:= r\left(a, s \bigm\vert A_i\right) = \frac{P\left(a \bigm\vert A_i\right)}{\underset{a'}{\max}\ P\left(a' \bigm\vert A_i\right)},
   \forall s \in \mathcal{S}.
\end{equation}

Moreover, since all triage actions are terminal,  only the reward participates in the target Q-value for triage actions:
\begin{equation}
   {\forall a \in \mathcal{A},\  Q_{\theta}^T\left(s, a \bigm\vert  e_i\right) = \vt{r}_a },
\end{equation} 

Consequently, to account for the counterfactual reward, we use a vector form of the temporal difference update where all actions participate in the error at each time step.

The reward for the action \textit{ask} is treated differently. As described in the next section, it is defined dynamically based on the quality of the current triage decision, to encode the notion that the agent should be efficient yet careful to gather enough information.

\subsection{Dynamic Q-Learning}
\label{dynamic_qn}
One key difference with the classic Q-Learning approach is the dynamic nature of ${Q_{\theta}^T\left(s, ask \bigm\vert e_i \right)}$, the target Q-value for the action \textit{ask}, which depends on the current Q-values of the triage actions. This dynamic dependency is especially useful given that the stopping and the triage part of the Dynamic Q-Learning (DyQN) agent are learning at the same time, and the value of asking for more information might change as the agent gets better at triage.  The ideal stopping criterion would stop the agent as soon as its highest Q-value corresponds to a correct triage decision, and do so reliably over all the vignettes. Assuming that the Q-values for the triage decisions are a good estimate of the probability of a particular triage, the DyQN approach is a heuristic which allows the agent to learn when best to stop asking questions given its current belief over the triage decisions. We develop two such heuristics in the form of probabilistic queries.

 \subsubsection{The OR query} 
 \label{ssec:or_query}
 The OR query is used by the \textsc{DyQN: or query} agent, as well as by the baseline agent \textsc{partially-observed: or query}. In practice, during each optimisation cycle and for each sampled memory $e_i$ in the batch, the Q-values for the starting state $s$ and following state $s'$ are computed. Given the parameters $\theta$ of the neural network, for state $s$, we refer to the maximum Q-value for triage actions as:
 \begin{equation}
 \begin{aligned}
     Q_m\left(s\right) & = \underset{a \in \mathcal{A}}{\max}\ Q_{\theta}\left(s, a\right),\\
 \end{aligned}
 \end{equation}
 and define ${\overline{Q}_m\left(s\right) = 1 - Q_m\left(s\right)}$. We then define the target Q-value for asking as:
\begin{equation}
\label{eq:target-q-asking-or}
Q_{\theta}^T\left(s, ask \bigm\vert e_i\right)  =  \overline{Q}_m\left(s\right)\ + Q_m(s)Q_m(s').
\end{equation}

 We see that this definition can be loosely mapped to the classic target Q-value, if one considers ${r\left(s, ask\right) = \overline{Q}_m\left(s\right)}$ and $\gamma = Q_m(s)$.  
 
To understand the origin of Eq. \eqref{eq:target-q-asking-or}, we must treat Q-values as probabilities and define the events $T$ and $T'$ as ``\emph{the agent's choice is an appropriate triage}'' on the current state $s$ and next state $s'$ respectively. Writing the event $\overline{T}$ as the negation of $T$, we define the probability of asking as:
\begin{equation}
    Q_{\theta}^T\left(s, ask \bigm\vert e_i\right) := P\left(ask \bigm\vert s\right) = P\left( \overline{T} \lor T' \bigm\vert s, s' \right),
\end{equation}
 that is the probability of the event ``\emph{Either the triage decision is not appropriate in the current state, \underline{or} it is appropriate in the next state}''. The query can also be written as:
\begin{equation}
    P\left( \overline{T} \lor T' \bigm\vert s, s' \right) = 1 -  P\left( T \land \overline{T}' \bigm\vert s, s' \right),
\end{equation}
which shows that the OR query encodes a stopping criterion heuristic corresponding to the event: ``\emph{the triage decision is appropriate on the current state, \underline{and} not appropriate on the next state}''.

 To recover equation \eqref{eq:target-q-asking-or}, we consider the Q-values for the triage actions as probabilities ${Q(s, a) = P\left(T \bigm\vert s, a\right)}$, then :
 \begin{equation}
 \begin{aligned}
 P\left(T \bigm\vert s\right) &= \sum_{a \in \mathcal{A}} \pi\left(a \bigm\vert s\right) P\left(T \bigm\vert s, a\right) \\ 
                   &= Q_m(s).
 \end{aligned}
 \end{equation}
Assuming the Markov property and ensuing conditional independencies ${\left( T \bigCI s', T'\ \bigm\vert\  s\right)}$ and ${\left(T' \bigCI s \bigm\vert s'\right)}$, we can write:
 \begin{equation}
 \begin{aligned}
 P\left( \overline{T} \lor T' \bigm\vert s, s' \right) &= P\left(\overline{T} \bigm\vert s\right) + P\left(T' \bigm\vert s'\right)\\
  &\phantom{=}\, -P\left( \overline{T} \land T' \bigm\vert s, s' \right)\\ 
  &= P\left(\overline{T} \bigm\vert s\right) + P\left(T' \bigm\vert s'\right)\\
  &\phantom{=}\, - P\left( \overline{T} \bigm\vert s \right)P\left( T' \bigm\vert s' \right)\\
  &= \overline{Q}_m\left(s\right) \ + Q_m(s') - Q_m(s') \left( 1 - Q_m(s) \right)\\
  &= \overline{Q}_m\left(s\right) \ + Q_m(s)Q_m(s').
 \end{aligned}
 \end{equation}

\subsubsection{The AND query} Among the other heuristics we tested, the AND query gave the best results regarding the stopping criterion, and is used by the \textsc{DyQN: and query} and the \textsc{partially-observed: and query} baseline.  For this query, we define the Q-value target for the \emph{ask} action as:
\begin{equation}
\label{eq:target-q-asking-and}
Q_{\theta}^T\left(s, ask \bigm\vert e_i\right) =  \overline{Q}_m\left(s\right) \left( Q_m(s') + \overline{Q}_m\left(s'\right)  Q_\theta(ask \bigm\vert s')\right).
\end{equation}

Contrary to the OR query, which can be viewed as a particular parametrisation of the reward and $\gamma$ of the classic Q-Learning target, the AND query has a form which is not immediately comparable.

In this case, the Q-target is obtained by considering the sequence of the event $T$ until the end of the interaction. That is, we consider the events $T_j$, $T_{j+1}$, ..., $T_k$, for states $s_j$ up to $s_k$, with $k$ the maximum number of questions. We then consider the probability $P_j$ of the event ``The current triage decision is incorrect, and the next is correct, \underline{or} both the current and next triage decision are incorrect, but the following triage decision is correct, \underline{or} \ldots and so on.''. We can rewrite $P_j$ as: 
 \begin{equation}
 \label{eq:p_and}
 \begin{aligned}
   P_j &= P\left( \bigvee\limits_{m=j+1}^k \left[ T_{m} \bigwedge\limits_{n=j}^{m-1} \overline{T}_n \right] \bigm\vert s_j, s_{j+1}  \right)\\ 
   &= P\left( \overline{T}_j \land T_{j+1} \bigm\vert s_j, s_{j+1}  \right) \\
   &\phantom{=}\, + P\left( \overline{T}_j \bigm\vert s_j  \right) P\left( \bigvee\limits_{m=j+2}^k \left[ T_{m} \bigwedge\limits_{n=j+1}^{m-1} \overline{T}_n \right] \bigm\vert s_j, s_{j+1}  \right)
   \\&\phantom{=}\, -  \underbrace{P\left( \bigwedge\limits_{m=j+1}^k \left[ T_{m} \bigwedge\limits_{n=j}^{m-1} \overline{T}_n \right] \bigm\vert s_j, s_{j+1}  \right)}_{ =\ 0}\\
   &=  P\left( \overline{T}_j \bigm\vert s_j  \right) 
   \left(P\left( T_{j+1} \bigm\vert s_{j+1}  \right)
   + P\left( \overline{T}_{j+1} \bigm\vert s_{j+1}  \right) P_{j+1} \right)\\
   &= \overline{Q}_m\left(s_j\right) \left( Q_m\left(s_{j+1}\right) + \overline{Q}_m\left(s_{j+1}\right)  Q_\theta\left(ask \bigm\vert s_{j+1}\right)\right).
\end{aligned}
\end{equation}

In practice, for both AND and OR queries, we obtained better results by using the known appropriate triages $A^{e_i}$ associated to the vignette of each sampled memories $e_i$ and defining $${Q_m\left(s\right) := Q_m\left(s \bigm\vert e_i\right) = \underset{a \in A^{e_i}}{max}\ Q(s, a)},$$that is the maximum Q-value associated with an appropriate triage.

\subsection{Memory}
The agent's memory is inspired from PER but does not rely on importance weighting. Instead, we associate to each memory tuple $e_{i} := (s,a,\vt{r},s')$ a priority: $$\nu_{i} = \left\vert \frac{1}{\vert \mathcal{A} \vert}\sum\limits_{a \in \mathcal{A}}  Q_{\theta}^T\left(s, a \bigm\vert e_{i} \right) - Q_{\theta}(s, a) \right\vert,$$
which relies on the vector form of the \emph{counterfactual reward} and is equal to the absolute value of the mean TD-Error over every action. We then store the experience tuple $e_i$ along with its priority $\nu_i$, which determines in which of the four \emph{priority buckets} the memory should be stored. The four priority buckets have different sampling probabilities, from $0.01$ for the lowest probability bucket to $0.8$ for the highest. 

Before each optimisation step, we sample $200$ experience tuples from the priority buckets, and every time a tuple is sampled, its priority decays with a factor ${\lambda = 0.999}$ which slowly displaces it into priority buckets with lower sampling probability. This approach contrasts with the priority update of PER, which sets the priority equal to the new TD-Error computed during the optimisation cycle. This approach yields better empirical results than using the classic PER priority update and importance weighting.

\subsection{Metrics}
\label{metrics}
We evaluated the quality of the model on a test set composed of previously unseen vignettes, using three target metrics: appropriateness, safety, and the average number of questions asked. During training, those metrics are evaluated over a sliding window of  20 vignettes, and during testing, they are evaluated over the whole test set. 

\paragraph{Appropriateness} 
Given a bag of triage decisions $A_i$, we define a triage $a$ as \emph{appropriate} if it lies at or between the most urgent $U(A_i)$ and the least urgent $u(A_i)$ triage decision for each vignette. For instance, if a vignette has two ground truth triage decisions  \{\textit{Red, Green}\} from two different doctors, the appropriate triage decisions are  \{\textit{Red, Yellow, Green}\}. Appropriateness is the ratio of agent's triage decisions which were appropriate over a set of vignettes. 

\paragraph{Safety} 
We consider a triage decision as safe if it lies at or above $u(A_i)$, the least urgent triage decision in $A_i$. Correspondingly, we define safety as the ratio of the agent's triage decisions which were safe over a set of vignettes. 

\paragraph{Average number of questions} The RL agent is trained to decide when best to stop and make a triage decision. The average number of questions is taken over a set of vignettes and varies between $0$ and $23$, an arbitrary limit at which point the agent is forced to make a triage decision. 

\begin{figure*}[!htb]
    \centering
    \includegraphics{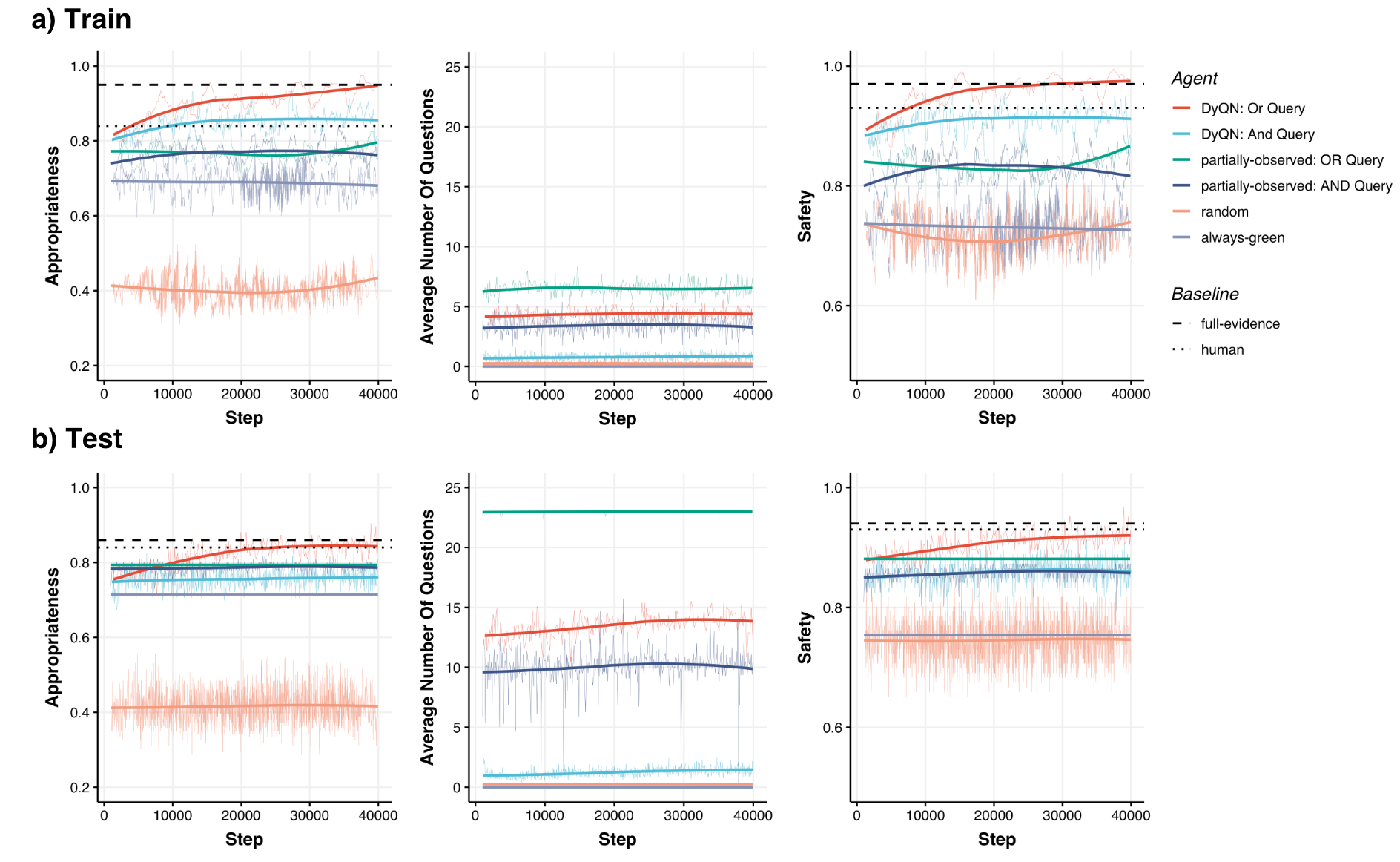}
    \caption{RL learning curves for all agents on the three key metrics show that, compared to the other learning agents, DyQN produces optimised policies which better match doctors' decisions while asking fewer questions on both the training and test set. Each step is a question asked, and after a burn-in period of 1000 steps, each new step is followed by a batched optimisation cycle. The lighter curves are the exponential mean average of the raw metric, and the darker curves are obtained using locally weighted smoothing. The higher variance of the training curves is due to a smaller sample size with a moving average of 20 training vignettes, as compared to the test curves where each record is the average over the full test set ($N = 126$).}
    \label{fig:learning_rate}
\end{figure*}

\subsection{Baselines}

\label{baselines}

We compare our RL approach with a series of baselines, using the same train ($N=1248$) and test split $(N=126)$ of the dataset $\mathcal{D}$. The supervised models are voting ensembles of classifiers calibrated using isotonic regression (see Table \ref{tab:annex_supervised_model}, and \ref{appendix:calibrated_versus_non_calibrated}). 

\paragraph{The \textsc{fully-observed} model} 
The \textsc{fully-observed} model is trained using the vignettes with their complete set of evidence $V_i$. It represents the less optimised version of the triage policy, which can only deal with full presentations. 

\paragraph{The \textsc{partially-observed} model} In addition to the two DyQn agents (\textsc{OR query} and \textsc{AND query}) defined above, we consider other two agents, referred to as \textsc{partially-observed} agents. The  \emph{learning agents} refers to the two DyQN agents and the two \textsc{partially-observed} agents, because those four agents learn to stop during the RL training. But the triage actions of the \textsc{partially-observed} agents are pre-trained in a fully supervised way on a greatly \emph{expanded dataset} of clinical vignettes $\mathcal{D}_{pow}$, constructed from the original set of vignettes $\mathcal{D}$. Given a vignette $V_i \in \mathcal{D}$, we generate a new vignette for each element of the powerset $\mathbb{P}\left(V_i\right)$ of the evidence set with a cap at $2^{10}$. If the vignette has more than ten pieces of evidence, the $k = \vert  V_i \vert - 10$ remaining evidences generate $k$ vignettes for each of the element of the powerset, by growing the rest of the evidence linearly and combining it with each element  $\mathbf{v} \in \mathbb{P}\left(V_i\right)$. For instance, if $|V_i| = 12$, we sample two pieces of evidence $\left\{v_{m}, v_n\right\} \in V_i$, and for each $\mathbf{v} \in \mathbb{P}\left(V_i \setminus\left\{v_{m}, v_n\right\} \right)$ one vignette will be created with evidence set $\mathbf{v}$, another with $\mathbf{v} \bigcup \left\{v_{m}\right\}$, and a third one with $\mathbf{v} \bigcup \left\{v_{m}, v_n\right\}$. Using the described process, we generated ${max\left(1, |V_i|-10\right) \times 2^{min \left(\vert V_i \vert , 10 \right) }}$ new vignettes from each vignette $V_i$ belonging to the original dataset $\mathcal{D}$.
Critically, for each created vignette, the correct triage decisions are the same as the generating vignette. 

After having trained a classifier on this extended dataset, the RL agent uses the class probabilities returned by the classifier as Q-values for the triage actions. In other words, only the \emph{ask} action is trained during the RL phase. Hence the \textsc{partially-observed} agent does not improve on its ability to triage given a fixed set of evidence but uses the RL process to train a stopping criterion. We present two sub-types of \textsc{partially-observed} agents, the \textsc{partially-observed: or query} which uses the same Q-value target defined in Eq. \eqref{eq:target-q-asking-or}, and the \textsc{partially-observed: and query} which uses the AND query defined in Eq. \eqref{eq:target-q-asking-and}.

\paragraph{Non-learning agents} We also compare our approach to a \textsc{random} policy, which picks random actions, and \textsc{always-green} policies which always picks the triage action \emph{Green} which has the highest prior probability in the dataset ($.48$).

\paragraph{The \textsc{human} baseline} The human performance on the triage task with full evidence is estimated using a proxy metric called the \emph{sample mean inter-expert agreement} for appropriateness $H_a$ and safety $H_s$. For each vignette $V_i$ and each associated bag of expert decisions $A_i$, this metric is the sample mean of the ratio of the experts' decisions $a \in A_i$ which were appropriate, or safe, given the decisions $A_i \setminus a$ from the other experts. Here $a$ represents an element of multiplicity $1$ in the multiset, that is only one expert decision. We then define human appropriateness and safety as:
\begin{equation}
\begin{aligned} 
    H_a & = \EX_{V_i \sim \mathcal{D}} \left[
    \frac{1}{\vert A_i\vert} 
    \sum_{a \in A_i}  \left[u\left(A_i \setminus a\right) \leq a \leq U\left(A_i \setminus a\right)\right] \right] \\
    H_s & = \EX_{V_i \sim \mathcal{D}} \left[
    \frac{1}{\vert A_i\vert} 
    \sum_{a \in A_i}  \left[u\left(A_i \setminus a\right) \leq a \right] \right]
\end{aligned}
\end{equation}

\section{Results}
\subsection{The fully supervised approach performs on a par with humans on the test set}
The \textsc{fully-observed} baseline reaches an appropriateness of $.86$ and safety of $.94$ on the test set (see \ref{tab:performance_test}). This baseline is trained in a fully supervised way on the full evidence sets $E_i$ of each vignette.  Contrary to the RL agents, it does not learn to stop and is not trained to handle small evidence sets, like those encountered during the RL interactions. If we consider the \textsc{human} appropriateness ($.84$) and safety ($.93$) as a good estimate of experts' performance on the task, the supervised baseline performs slightly better than humans on full evidence sets. However, using only the supervised approach does not give us a direct insight into when best to stop asking questions, and if implemented on its own would require the definition of an arbitrary stopping criterion.  

\begin{figure}[ht]
\centering
\vspace{0pt}
\begin{tabular}{l c c r c}
\toprule
  & Appropriateness &  Safety & Avg. Questions & N\\ 
\cmidrule(lr){2-5} 
\textbf{\textsc{DyQN: or query}} & \textbf{.85} (.023) & .93 (.015) & 13.34 (.875) & 10 \\     
\textsc{DyQN: and query} & .76 (.014) & .86 (.012) & 1.40 (.331) & 10 \\     

\textsc{partially-observed: or query }& .79 (.000) & .88 (.000) & 23 (.000) & 10\\ 
\textsc{partially-observed: and query}& .79 (.006) & .86 (.005) & 10.35 (.467) & 10\\ 
\textsc{random} & .39 (.027) & .74 (.026) & .25 (.025) & 10\\    
\textsc{always-green} & .71 (.000) & .75 (.000) & 0 (.000) & 10\\ \cmidrule(lr){1-5} 
\textsc{human} & .84  & .93   & full &  \\
\textsc{fully-observed} & .86  & .94  & full &  \\
\bottomrule
\end{tabular}
\captionof{table}{Average performance on the test set at convergence (over the last 10 evaluations), after training the agents on a common training and test set. }
\label{tab:performance_test}
\end{figure} 

\subsection{DyQN performs on a par with both experts and the fully supervised approach}
To compare DyQN to the baselines, we trained all agents on the same training and test set (Figure \ref{fig:learning_rate}). The results presented in Table \ref{tab:performance_test} summarize the performance of the \emph{learning agents} with the best hyper-parameters: \textsc{DyQN: or query}, \textsc{DyQN: and query}, \textsc{partially-observed: or query }, \textsc{partially-observed: and query}. The DyQN agent using the OR query performs better than the other agents in term of appropriateness ($M = .85$, $CI95 = .023$, $min = .81$, $max = .90$) and safety ($M = .93$, $CI95 = .015$, $min = .90$, $max = .97$). And while it relies on less clinical evidence, asking on average $13.3$ questions ($CI95=.015$, $min = 10.8$, $max = 15.1$), it is on a par with \textsc{human} performance ($.84$ appropriateness) as well as the \textsc{fully-observed} baseline ($.86$ appropriateness), both of which use all the evidence on the vignette to come to a decision. Interestingly, the AND query produces comparatively worse result for the \textsc{DyQN} agent than it does for the \textsc{partially-observed} agent (see section \ref{section:partially_perf}).

\subsection{The agents adapt to unseen cases by asking more questions}

\begin{figure}[ht]
    \centering
    \resizebox{0.8\linewidth}{!}{
    \includegraphics{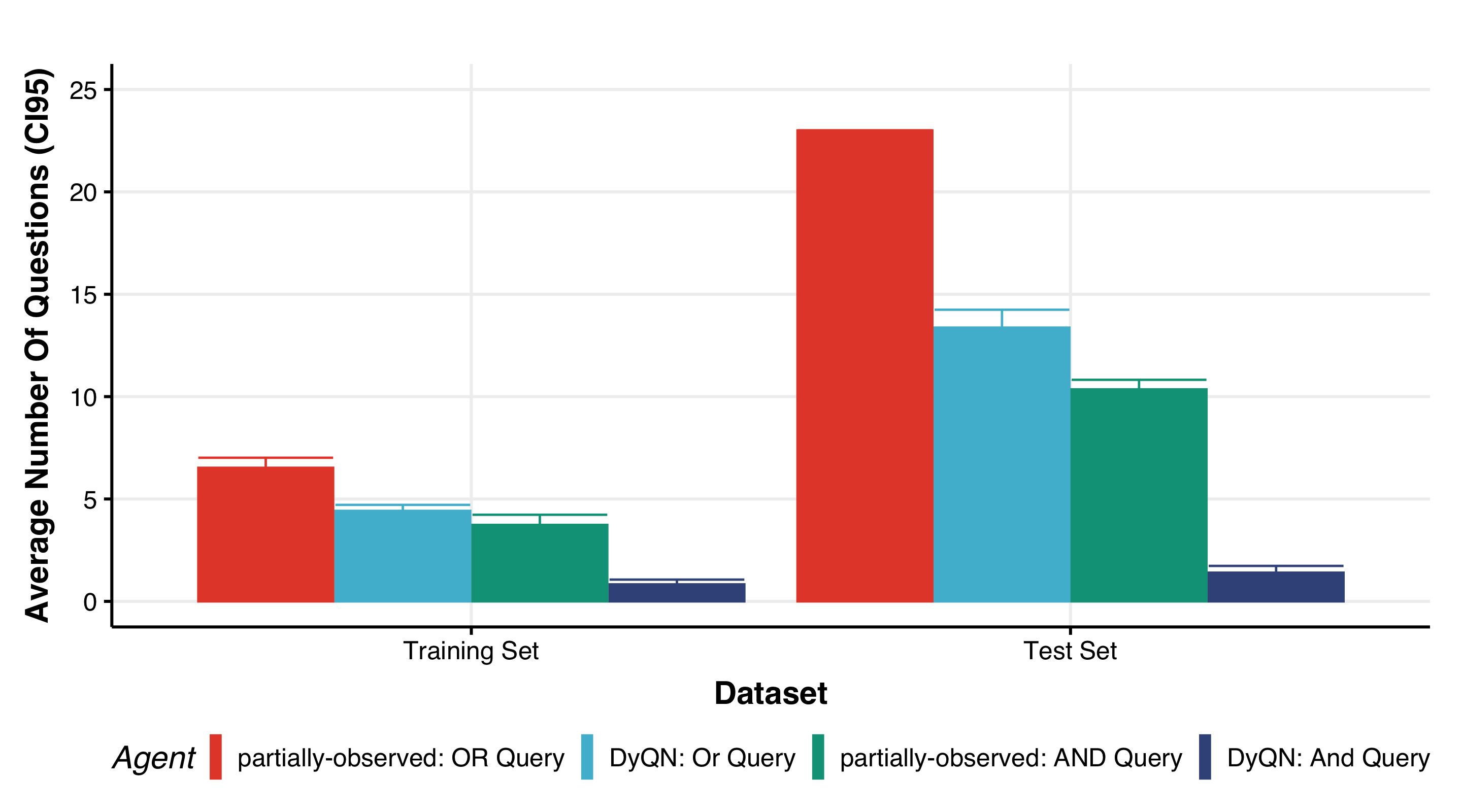}}
    \caption{The difference in number of questions asked between training and testing across the four learning agents shows that the Dynamic Q-value targets allows the agent to adapt to unseen cases by asking more questions.}
    \label{fig:bar_question_number}
\end{figure}

The average number of questions asked during training for all the learning agents is significantly lower ($M=4.9$, $CI95=.50$) than the number of questions asked during testing ($M=15.6$, $CI95=2.06$) (Figure \ref{fig:bar_question_number}). This is a direct effect of the dynamic nature of the target Q-value, which adapts to the agent's confidence in its triage decision. Given that the presentations on the test set are new to the agent, the \emph{ask} action will have a high Q-value, and be favoured over triage, until the agent gathers enough information to make a high confidence triage decision. 
\subsection{The \textsc{partially-observed} baseline performance}
\label{section:partially_perf}
The \textsc{partially-observed} agents yield similar results in terms of appropriateness ($M=.79$), with a slight increase in safety for the \textsc{partially-observed: or query} which obtains $88\%$ safety on average on the test set, compared with $84\%$ for \textsc{partially-observed: and query} (Table \ref{tab:performance_test}). The learning curves presented in Figure \ref{fig:learning_rate} show that the appropriateness and safety of the two \textsc{partially-observed} baselines do not improve significantly across training. This is due to their fixed triage model, pre-trained in a supervised manner on $\mathcal{D}_{pow}$, while only the \emph{ask} action is trained during RL. 

What differentiates the two agents is the number of questions they ask. \textsc{partially-observed: or query} asks on average more questions ($M=6.5$) than \textsc{partially-observed: and query} ($M=3.7$) on the training set, and never stops early on the test set ($M=23$ questions) while \textsc{partially-observed: and query} stops on average after $10.35$ questions. 

The nature of the OR query might not be suited for a model \emph{calibrated} to work well across all size of evidence sets, similar to the pre-trained model the two \textsc{partially-observed} agents use for triage. Indeed, for well-calibrated models, the quality of the triage decision tends to increase monotonically with the size of the evidence set (see \ref{appendix:calibrated_versus_non_calibrated}), which leads the OR query to favour the \emph{ask} action. Because the underlying heuristic of its stopping criterion is proportional to $P(T  \land \overline{T}' | s, s')$, and it tends to be triggered when the quality of the triage decision decreases as the evidence set grows (i.e. in non-calibrated models). 

The AND query, on the other hand, puts more emphasis on the quality of the current triage decision, and when $Q_m(s)$ is $1$ the target Q-value for \emph{ask} is $0$. The agent then becomes increasingly myopic as the current Q-values for the triage actions increase, and it will tend to discard the future Q-values. This property may aid \textsc{partially-observed: and query} to adapt to a well-calibrated model, assigning a higher weight to the present decision, and explain why it can stop on the test set. On the other hand, the same mechanism may impact the performance of the \textsc{DyQN: and query} agent by being too confident in the current decision and stopping early while the triage model is still early in its training. This quickly decreases the ability of the agent to explore, and in turn, increases the bias of the triage model towards smaller evidence sets.
 
\subsection{DyQN shows a slightly worse performance on K-Fold cross-validation}
The performance of the \textsc{DyQN: or query} agent was then evaluated using three different K-Fold cross-validation runs, with ten folds each. The results were slightly worse on average than the performance obtained on the test split used for the agent's comparisons.  While the performance varied greatly across runs (see \ref{fig:appendix_kfold_learning} in the appendix), on the last ten evaluations of the test set for each fold, the average appropriateness reached $0.81$ ($min=.68$, $max=.92$, $CI95=.004$, $N=271$), with a safety of $.90$ ($min=0.80$, $max=0.97$, $CI95=0.002$, $N=271$), and an average number of questions asked of 13.14 ($CI95=.105$). The variability in performances across runs is a classic observation in Reinforcement Learning, but here, it is also due to the diverse difficulty of the randomly sampled test vignettes.  The lower average appropriateness, which did not reach the average human performance ($.84$ evaluated over the whole dataset),  may indicate that on average the random test splits were more difficult, i.e. further from their respective training set distribution, than the one sampled for the agents' comparison. We did not submit the other agent to the same K-Fold cross-validation, but we would expect similar variability in the results as well as a lower performance overall.

\section{Discussion}

By learning when best to stop asking questions given a patient presentation, the DyQN is able to produce an optimised policy which reaches the same performance as supervised methods while requiring less evidence. It improves upon clinician policies by combining information from several experts for each of the clinical presentations. Moreover, while the result on the test set is on a par with human performance, the performance of the fully supervised approach on the training set  ($M = .95$ appropriateness) indicates that the task has a low Bayes Error rate, and given enough data we would expect DyQN to exceed human performance.  

One of the reasons to use the Dynamic Q-Learning over classic Q-Learning is to ensure that the Q-values correspond to a valid probability distribution. Using the classic DQN would produce unbounded Q-values for the asking action, because asking is not terminal, whereas the Q-values for the triage action would be bounded. In classic Q-Learning, only a careful process of reward shaping for the \emph{ask} action could account for this effect. 

While the problem of optimal stopping has been studied in settings where actions are associated with a cost \cite{Yu2007, DeFarias2000}, the other immediate advantage of DyQN is that it is able to treat the stopping heuristic as an inference task over the quality of the agent's triage decisions. Interpreting the triage actions' Q-values as probabilities allow us to rewrite the Q-value update as the solution to the inference query, which leads to the agent getting increasingly better at it through interaction, and adapting dynamically as triage decisions improve during training. This approach is well-tailored for information gathering tasks, where an agent must make inference on a latent variable (here the triage) given the information it has gathered so far. 

Our particular task sits at the intersection between a supervised task, which allows obtaining rich \emph{counterfactual rewards} at each step, and an RL task, which allows learning to triage and stop simultaneously. Contrary to a purely supervised approach, the stopping criterion impacts the triage decision, and vice-versa, and both systems learn jointly to optimise the policy.  Indeed, it is the nature of the RL process to bias the collection of data towards trajectories associated with high values.  In our case, the value of the \emph{ask} action depends on the triage decisions, but because stopping impacts the data collected, it affects, in turn, the quality of the triage decisions. While exploration is critical for RL agents, they do not explore the state-space exhaustively, which is an advantage in very large state-spaces like ours. This might explain in part why the DyQN is able to outperform the  \textsc{partially-observed} baseline performance on the training and test sets, because the joint optimisation of stopping and triage produces a data distribution which favours the DyQN triage performance. On the other hand, this data gathering process may also lead to significant biases due to important regions of the state-space left entirely unexplored.

This approach gives promising results for the future of data-driven triage automation and could allow learning region-specific policies or secondary triage policies for which no guidelines are available. However, like any medical algorithm, it requires thorough clinical validation, before and after deployment, in order to ensure its safety and efficacy. Furthermore, in practice, given the breadth and complexity of clinical decision making, these new expert-systems based on machine-learning are often associated with a rule-based layer which guarantees that corner cases are covered.

\section{Conclusion}
In this work, we introduce a method to learn medical triage from expert decisions based on Dynamic Q-Learning, a variant of Deep Q-Learning, which allows a Reinforcement Learning agent to learn when to stop asking questions by learning to infer the quality of its triage decision. Our approach can be used in conjunction with any question-asking system, and while requiring less evidence to come to a decision, the best DyQN agent is on a par with experts' performance, as well as with a fully supervised approach. This RL approach can produce triage policies tailored to healthcare settings with specific triage needs. Moreover, it could help improve clinical decision making in regions where trained experts are scarce. A direction for future investigation should be to train agents not only to stop but also to learn a policy over questions under an active-inference framework.

\section*{Conflict of Interest Statement}
 The Chief Investigator and most co-investigators are paid employees of Babylon Health. 

\section*{Author Contributions}
A.B. conceived of the presented idea and developed the theory. A.B. developed the software necessary for training reinforcement learning agents and ran the experiments presented in the paper. A.B. supervised the work of B.B., Y.Z. and M.L. who performed the initial experiments. B.B. developed and tested many of the theories leading to the final algorithm. Y.P. and A.Ba. participated in the formal definition and gathering of the dataset. G.P., R.B., K.G., D.T., J.R., A.Ba., Y.P., Y.Z. and D.B. verified the analytical methods.  D.B. and S.J. helped supervise the project. All authors discussed the results and contributed to the final manuscript. 

\section*{Acknowledgments}
The authors would like to thank Mario Bordbar, Nathalie Bradley-Schmieg,  Lucy Kay, and Karolina Maximova, for their unwavering support.

\bibliographystyle{unsrt}

\appendix
\section{Baseline model for supervised triage}
We train the supervised model using a  soft-voting ensemble classifier from \texttt{scikit-learn} \cite{PedregosaFABIANPEDREGOSA2011}, calibrated using a non-parametric approach based on isotonic regression (see \emph{sklearn.isotonic}). The structure of the ensemble is described in table \ref{tab:annex_supervised_model}.

\begin{figure}[ht] 
\centering

\begin{tabular}{l l}
\toprule
Type  &  Parameters \\ 
  &  \\ 
\cmidrule(lr){1-2} 
SGDClassifier & max\_iter=1000 \\
\cmidrule(lr){1-2} 
LogisticRegression & max\_iter=1000 \\
\cmidrule(lr){1-2} 
MLPClassifier & hidden\_layer\_sizes=(512, 512),\\
& alpha=1, max\_iter=1000, \\
& n\_iter\_no\_change=5, tol=0.001 \\
\cmidrule(lr){1-2} 
DecisionTreeClassifier & max\_depth=5 \\
\cmidrule(lr){1-2} 
RandomForestClassifier & max\_depth=5, n\_estimators=10,\\ & max\_features=1 \\
\cmidrule(lr){1-2} 
SVC & gamma=``auto'' \\
\bottomrule
\end{tabular}
\captionof{table}{Structure of the supervised triage model. Each  estimator of the ensemble was calibrated using the \emph{isotonic} method of \emph{CalibratedClassifierCV}.}
\label{tab:annex_supervised_model} 
\end{figure}

\section{Non-Calibrated versus Calibrated voting classifier}
\label{appendix:calibrated_versus_non_calibrated}
\begin{figure*}[!htb]
    \centering
    \includegraphics{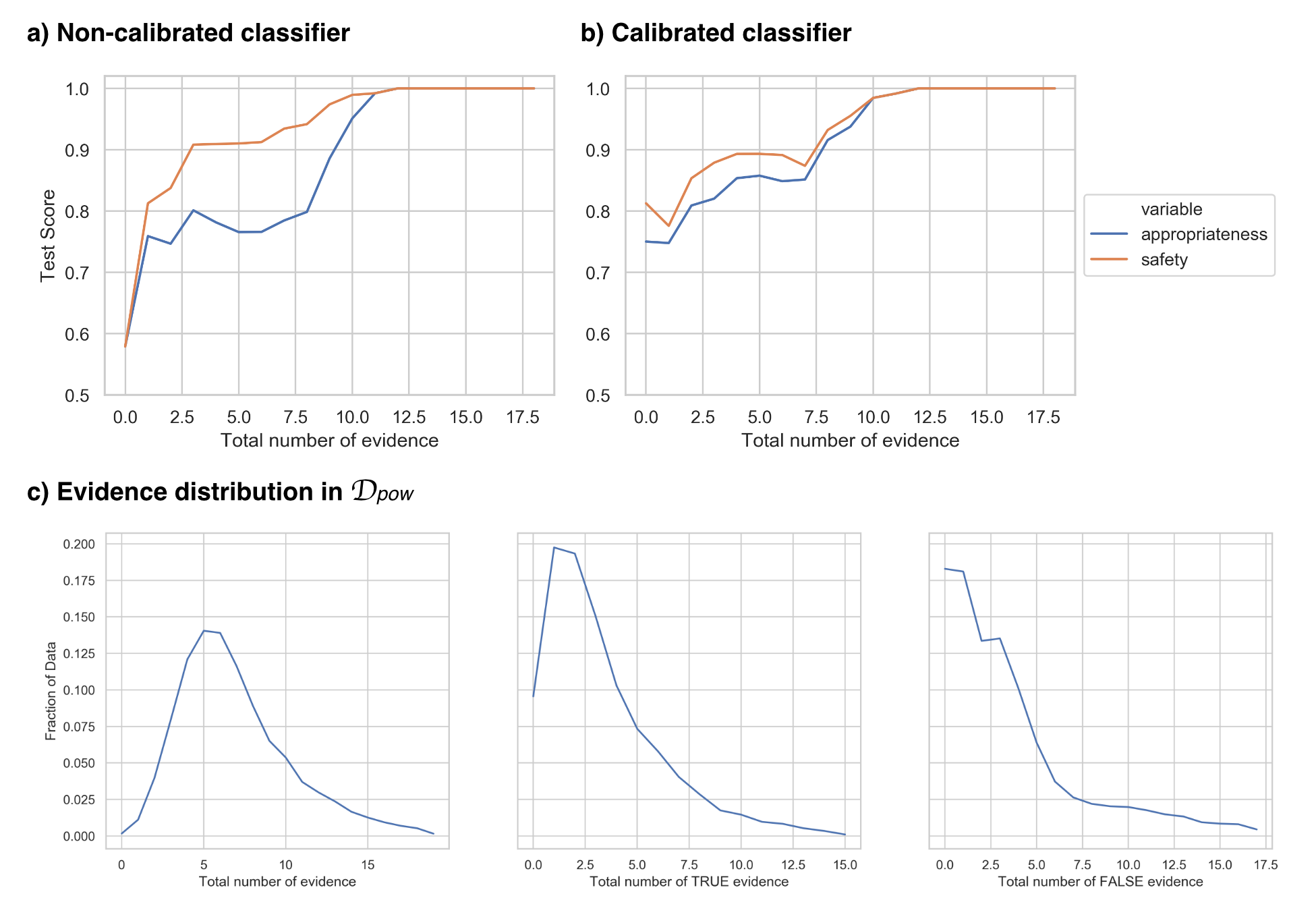}
    \caption{Performance comparison of the supervised approaches, trained with a) and without b) isotonic calibration. c) presents the distribution of the evidence in the test set used for the comparison, a random sample of the $\mathcal{D}_{pow}$ test set ($N=9595$).}
    \label{fig:annex_calibrated_vs_non_calibrated}
\end{figure*}

Calibrating the voting classifier for both the partially-observed and fully-observed baselines results in better performance on smaller evidence set. The performance is also more consistent overall, and tend to improve monotonically as the evidence set grows. Figure \ref{fig:annex_calibrated_vs_non_calibrated} shows the performance of both approaches over a random sample of 10\% of the $\mathcal{D}_{pow}$ test set ($N=9595$),

\section{DyQN K-Fold cross-validation}
\begin{figure*}[!htb]
    \centering
    \includegraphics{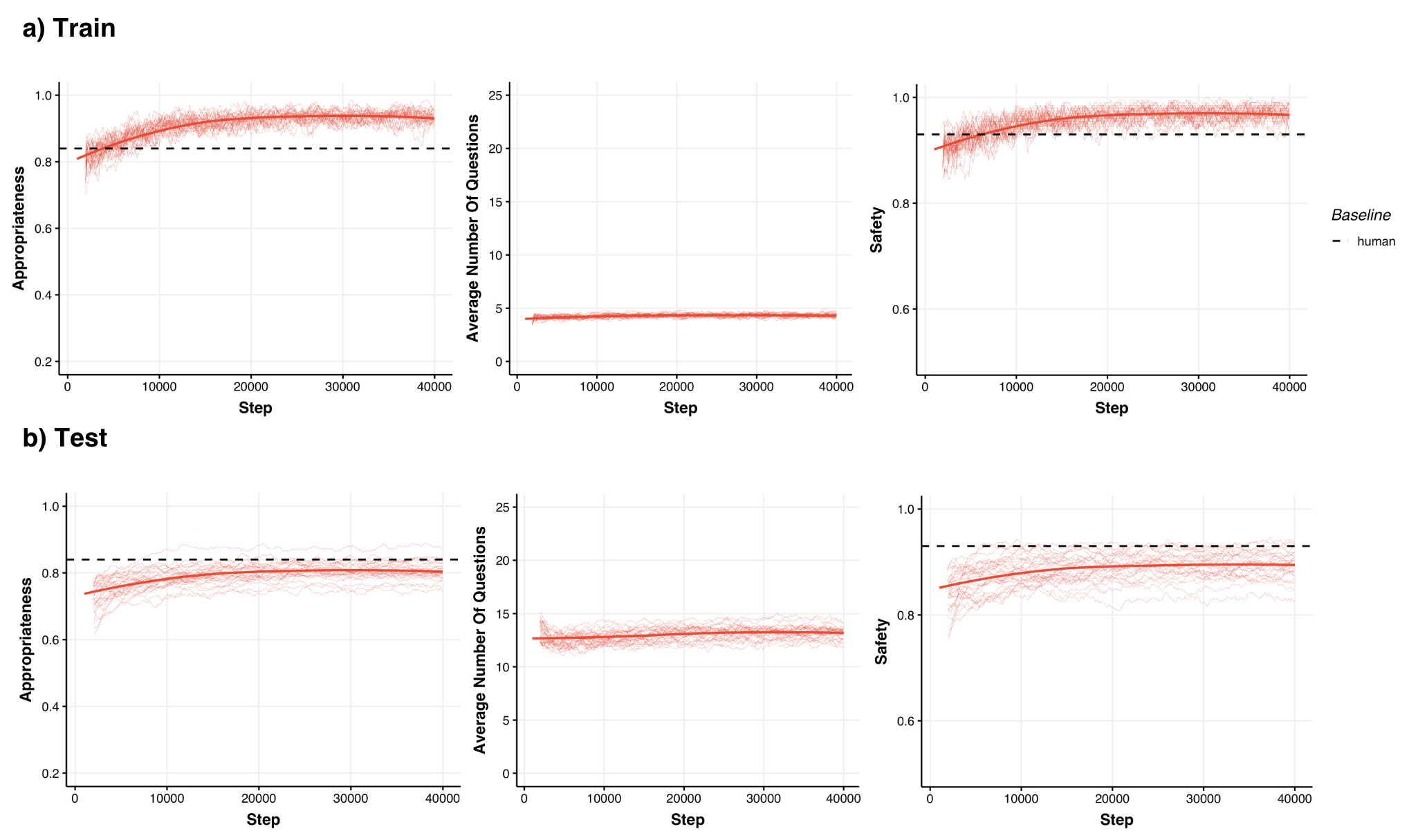}
    \caption{\textsc{DyQN: or query} learning curves during $\left(3 \times\right)$ 10-Fold cross-validation.}
    \label{fig:appendix_kfold_learning}
\end{figure*}

The learning curves of the \textsc{DyQN: or query} agent across three different 10-Fold cross-validation runs, started with three different random seeds, are presented figure \ref{fig:appendix_kfold_learning}.

\begin{algorithm*}
    \caption{Training cycle of the Dynamic Q-Network (DyQN)}
    \label{algo:dyqn}
  \begin{algorithmic}[1]
    \REQUIRE DyQN's $Q_{\theta}(s, a)$ and $Q^T_{\theta}\left(s, a \bigm\vert e \right)$ functions, environment's $step(a)$,
    memory's $store(s,a,r,s',\nu)$ and $sample(size)$, noise variance $\sigma(t)$. 
    \INPUT  dataset $\mathcal{D}$ of clinical vignettes 
    \STATE \textbf{Initialization} $\theta \gets \theta_0$
    \FOR{$i \gets 1$ to $N$}
    \COMMENT{until the maximum number of games is reached.}
      \STATE $V_i \sim \mathcal{D}$
      \STATE $s_0 = step(\varnothing)$
      \STATE $stop = False$
      \FOR{$k \gets 0$ to $K$}
        \COMMENT{until maximum question is reached.}
        \IF{$stop$} \COMMENT{the environment forced a triage action.}
            \STATE $A = \mathcal{A} $
        \ELSE
            \STATE $A = \mathcal{A}^+ = \mathcal{A} \bigcup ask$
        \ENDIF
        \STATE ${a_k = \underset{a \in A}{\argmax}\ Q_{\theta}(s_k, a) + \left[a = ask\right] \mathcal{N}\left(0, \sigma\left(i\right)\right)}$  \COMMENT{noise is added to the Q-value for \emph{ask} before greedy selection.}
        \STATE ${s_{k+1},\ \vt{r}_k,\ stop = step(a_k)}$
        \STATE $e_k = \left(s_k, a_k, \vt{r}_k, s_{k+1}\right)$
        \STATE $\nu_k =  \left\vert \frac{1}{\vert \mathcal{A} \vert}\sum\limits_{a \in \mathcal{A}}  Q^T_{\theta}\left(s_k, a \bigm\vert e_k \right) - Q_{\theta}(s_k, a) \right\vert$ \COMMENT{compute memory priority.}
        \STATE ${store(e_k, \nu_k)}$
        \IF{$i \geq L$} 
            \COMMENT{after the burn-in period perform one optimization cycle at each step.}
            \STATE $\xi = sample(N)$
            \COMMENT{Sample a batch of size N from memory.} 
             \STATE ${\mathcal{L}(\theta) = \frac{1}{N}\sum\limits_{e_i \in \xi}\ \sum\limits_{a \in \mathcal{A}^+} \left( Q^{T}_{\theta}\left(s, a \bigm\vert e_i \right) - Q_{\theta}\left(s, a\right) \right)^2}$
             \STATE $\theta \gets \theta - \alpha \Delta_{\theta} \mathcal{L}(\theta)$
        \ENDIF
        \IF{$a_k \neq ask$} \COMMENT{sample new vignette when a triage decision is made.}
            \BREAK
        \ENDIF
           
      \ENDFOR
    \ENDFOR
  \end{algorithmic}
\end{algorithm*}

\end{document}